\documentclass{article}

\usepackage{microtype}
\usepackage{graphicx}
\usepackage{subfigure}
\usepackage{booktabs} 

\usepackage{tikz}
\usepackage{bm}
\usepackage{color}
\usepackage{amsmath}
\usepackage{amssymb}

\usepackage{hyperref}

\usepackage[accepted]{icml2019}

\icmltitlerunning{Rare Disease Detection by Sequence Modeling}

\begin{document}

\twocolumn[
\icmltitle{Rare Disease Detection by Sequence Modeling with Generative Adversarial Networks}



\icmlsetsymbol{equal}{*}

\begin{icmlauthorlist}
\icmlauthor{Kezi Yu}{iqvia}
\icmlauthor{Yunlong Wang}{iqvia}
\icmlauthor{Yong Cai}{iqvia}
\icmlauthor{Cao Xiao}{iqvia}
\icmlauthor{Emily Zhao}{iqvia}
\icmlauthor{Lucas Glass}{iqvia}
\icmlauthor{Jimeng Sun}{gatech}

\end{icmlauthorlist}

\icmlaffiliation{iqvia}{IQVIA Inc., Plymouth Meeting, PA, USA}
\icmlaffiliation{gatech}{Georgia Institute of Technology, Atlanta, GA, USA}

\icmlcorrespondingauthor{Kezi Yu}{kezi.yu1@iqvia.com}

\icmlkeywords{Rare Disease Detection, Generative Adversarial Networks, Recurrent Neural Networks}

\vskip 0.3in
]



\printAffiliationsAndNotice{}  

\begin{abstract}
Rare diseases affecting 350 million individuals are commonly associated with delay in diagnosis or misdiagnosis. To improve those patients' outcome, rare disease detection is an important task for identifying patients with rare conditions based on longitudinal medical claims. In this paper, we present a deep learning method for detecting patients with exocrine pancreatic insufficiency (EPI) (a rare disease). The contribution includes  1) a large longitudinal study using 7 years medical claims from 1.8 million patients including 29,149 EPI patients, 
2) a new deep learning model using generative adversarial networks (GANs) to boost rare disease class, and also leveraging recurrent neural networks to model patient sequence data, 
3) an accurate prediction with 0.56 PR-AUC which outperformed benchmark models in terms of precision and recall. 
\end{abstract}

\section{Introduction}
\label{sec:intro}
Rare disease affect 350 million patients worldwide \cite{kaplan2013priority}. Collectively they are common but individually they are rare. Given rare diseases' low prevalence rate among population, the low disease awareness could lead to patients being misdiagnosed/undiagnosed and not getting the appropriate treatment. Patients with rare diseases often visit several physicians over the course of many years before they receive diagnoses for their conditions~\cite{boat2011rare}. An effective detection method is crucial to help raise disease awareness and achieve early disease intervention~\cite{cameron2010evaluation}. On the other hand, interest in machine learning for healthcare has grown immensely during last several years~\cite{ching2018opportunities}. Several machine learning methods, such as Recurrent Network~\cite{choi2016using, lipton2015learning}, auto encoder~\cite{miotto2016deep}, FHIR-formatted representation~\cite{rajkomar2018scalable}, etc. have been proposed to predict patient-level disease using electronic healthcare record (EHR) data. For more comprehensive overview of machine learning application on healthcare, we refer readers to~\cite{xiao2018opportunities, ghassemi2018opportunities, obermeyer2016predicting}. 

Recently, deep learning based models, such as long short-term memory and attention models, have been widely applied for disease detection and made improvements on prediction accuracy. In  \cite{choi2016doctor}, the authors proposed an approach for converting the patient history into medical sequence and then train a long short term memory for sequence labeling task, based on which, an application was developed in \cite{choi2016using}. To enhance the interpretability, there have been great efforts of trying to explain black-box deep models, including via attention mechanism~\citep{choi2016retain, choi2017gram}, decay factor\ cite{bai2018interpretable}, mimic decisions of deep models with decision tree~\citep{Che2016InterpretableDM,che2017rnn}, etc.

Generative adversarial networks (GANs) \cite{goodfellow2014generative} have drawn numerous attention for its potential to generate almost true samples from random noise inputs. Although the original idea was more focused on the generator, in \cite{salimans2016improved, dai2017good}, the authors proposed to use GANs in a semi-supervised learning (SSL) setting and demonstrated that GANs performed well by leveraging unlabeled data with novel training techniques. Meanwhile, the problem of rare disease detection falls perfectly under the setting of semi-supervised learning. Since the diagnosed patient population is extremely small, we have limited positive samples but a large number of patients who are under-diagnosed. 

Instead of using hand-crafted features as input to a classifier \cite{li2018semi}, we directly worked with patient medical history sequences. This comes with multiple benefits, including the ability to capture more complex disease patterns, saving extensive efforts in feature engineering,  and making the framework easier transferable to another disease. Since GANs were not intrinsically able to handle sequence data, we opted to use recurrent neural network (RNN)-based model for fix-length sequence embedding. 

\section{Method}
\begin{figure}[h!]
    \centering
    \includegraphics[width=.45\textwidth]{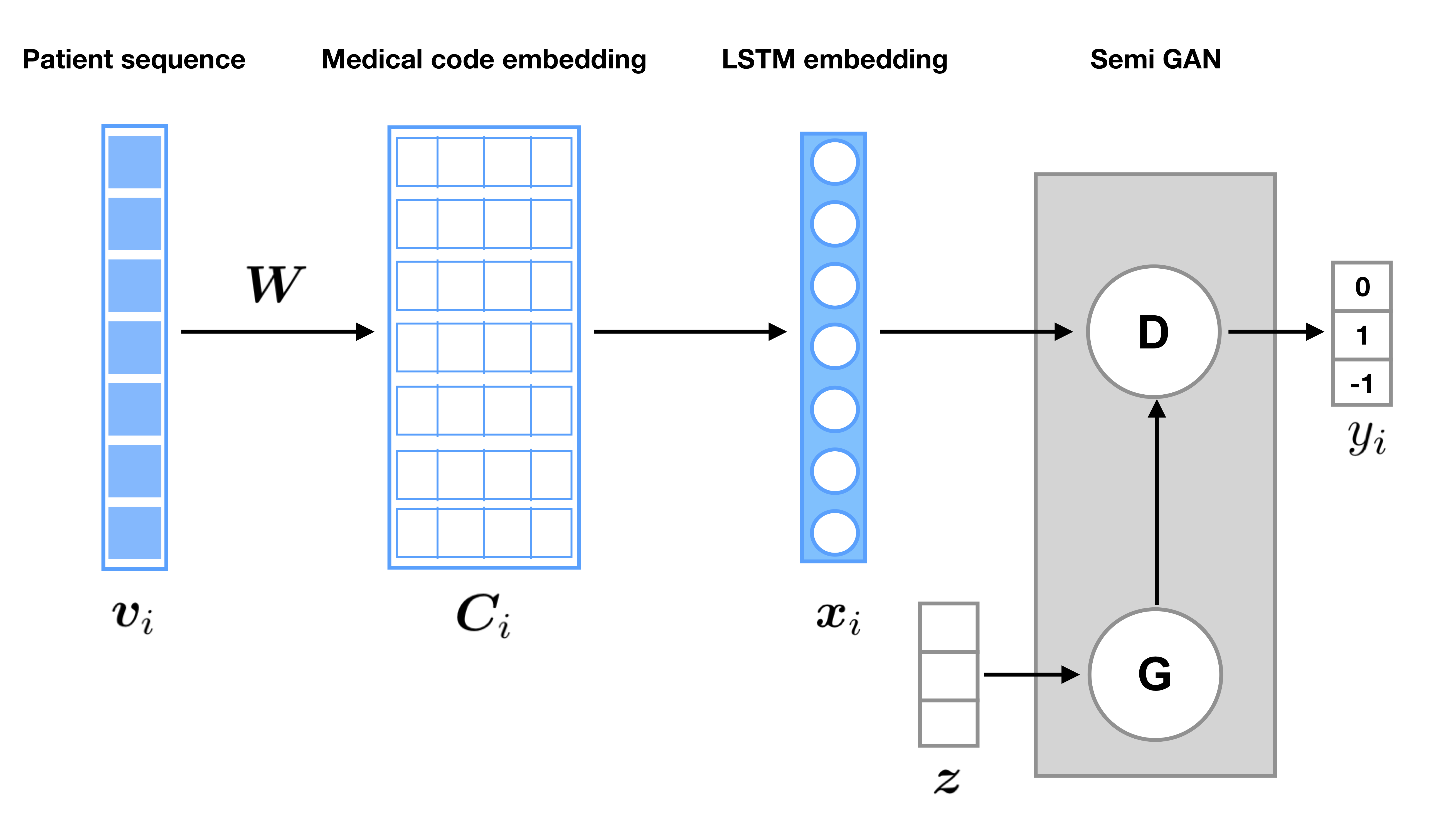}
    \caption{Framework architecture illustrated. $z$ is a random noise input to the generator of GAN.}
    \label{fig:arc}
\end{figure}

The architecture of our framework is shown in Figure \ref{fig:arc}.
\vspace{-3pt}
\vspace{-3pt}

\begin{figure}[h]
\centerline{

\def \thisplotscale {1}
\def \unit {\thisplotscale cm}

\pgfdeclarelayer{back}
\pgfdeclarelayer{fore}
\pgfdeclarelayer{mid}
\pgfsetlayers{back,mid,fore}

\begin{tikzpicture}[-stealth,  shorten >=0, x = 0.98*\unit, y=0.6*\unit]

\begin{pgfonlayer}{fore}
    \node at (0, 0) {$Dx_1$};
    \node at (1, 0) {$Dx_2$};
    \node at (2, 0) {$Rx_1$};
    \node at (3, 0) {$Dx_1$};
    \node at (4, 0) {$Rx_2$};
    \node at (5, 0) {$Px_1$};
    \node at (6, 0) {$Px_2$};
    \node at (7, 0) {$Rx_3$};
    
    \draw[->] (-0.5, -1.1) -- (7.5, -1.1);
    \node at (8, -1.2) {time};
    

\end{pgfonlayer}
    
\begin{pgfonlayer}{mid}
    \fill[blue!40!white] (-0.4, 0.4) rectangle (0.5, -0.4);
    \fill[blue!40!white] (0.6, 0.4) rectangle (1.5, -0.4);
    \fill[green!40!white] (1.6, 0.4) rectangle (2.5, -0.4);
    \fill[blue!40!white] (2.6, 0.4) rectangle (3.5, -0.4);
    \fill[green!40!white] (3.6, 0.4) rectangle (4.5, -0.4);
    \fill[red!40!white] (4.6, 0.4) rectangle (5.5, -0.4);
    \fill[red!40!white] (5.6, 0.4) rectangle (6.5, -0.4);
    \fill[green!40!white] (6.6, 0.4) rectangle (7.5, -0.4);
\end{pgfonlayer}
     	
\end{tikzpicture} }
\caption{A toy example of a patient medical history sequence. Dx refers to diagnosis, Rx refers to prescription and Px refers to medical procedure. The subscripts denote different codes within each category.} 
\label{fig:pat_seq}
\vspace{-1pt}
\end{figure}
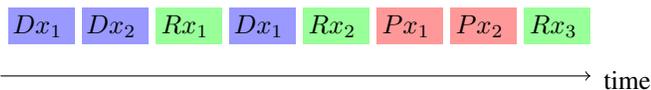

Each patient is represented by a sequence $\bm{v}_i = \{v_{ij}, j=1,\dots, N\}$, of which $v_{ij}$ is a medical code indicating a type of hospital visit (Dx) or prescription (Rx). A graphical illustration of such representation is shown in Figure \ref{fig:pat_seq}. The patient sequence is then transformed to its matrix representation $\bm{C}_i$ by embedding the medical codes, i.e. $\bm{c}_{ij} = \bm{W}h(v_{ij})$, where $\bm{W}$ is embedding matrix and $h(\cdot)$ denotes one-hot encoding. Then an LSTM network is used to encode the sequence to $\bm{x}_i$. The embedded medical sequence is fed into the discriminator $D$ of a SSL GAN, where the prediction is either positive (1), negative (0) or generated sample by generator $G$ (-1).

\subsection{Patient Record Embedding}
\label{subsec:mce}

\textbf{Encode medical codes.}
In the patient medical history sequence, each medical code is essentially a categorical variable. The number of categories (different types of medical codes) depends on the level of specificity, i.e., the more specific the meaning of a code is, the more unique codes there would be. In our case, we end up with 5362 unique codes. 

We are inspired by the concept of word embedding in natural language processing (NLP) community. Some notable examples include word2vec \cite{mikolov2013distributed} and Glove \cite{pennington2014glove}. In the original application, the vector representations retain semantic meanings, e.g., synonyms of a word tend to be closer to each other spatially. We will demonstrate a similar behavior of our medical code embedding in a later section.

We used skip-gram model with negative sampling to train the embedding network. The minimum count for valid code is 5, i.e., any codes that occur less than five times in all the sequences would be discarded for training the embedding model, and they were assigned a all-zero vector as their embedding. This left us with 5035, or 93\% of all the unique codes. The dimension $d_{w}$ of embedding vector was empirically chosen to be 300. 

\noindent\textbf{Embed Longitudinal Records.}
The most prominent model for processing time series or sequence data is recurrent neural network (RNN). Different variations of memory cells, including long short-term memory (LSTM) \cite{hochreiter1997long} and gated recurrent unit (GRU) \cite{chung2014empirical}, were proposed to handle the long-term dependencies over time, which significantly improved the performance of RNNs in various types of tasks, such as sequence classification, sequence tagging \cite{huang2015bidirectional}, and machine translation \cite{cho2014properties}. A commonly used technique in various tasks is to use the hidden state vector as a representation of the sequence. 

We adopted the same idea for sequence embedding. Specifically, patient sequences were padded with a fixed length of $N$. Only labeled training patient sequences were used for training LSTM embedding model. A single-layer LSTM with dimension of the hidden state equal to $d_S$ was used. The hidden state of each time stamp was retained, and then aggregated by max pooling operation over time, which resulted in a $d_S$-dimension vector. In our experiments, we empirically chose $N=300$ and $d_S=256$. After sequence embedding, we appended the patients' age and gender, and then scale all the features to the range between -1 and 1. The final feature vector has a dimension of 258.

\subsection{Semi-supervised GAN}
In the original framework of GANs \cite{goodfellow2014generative}, a GAN has a generator network $G$ that takes random noise as input and produce samples that follow the real data distribution $p_{data}(x)$. The training of $G$ is guided by the discriminator network $D$, which is trained to distinguish samples from the generator distribution $p_{model}(x)$ from real data. Suppose that the goal of $D$ extends to finding the actual class assignment of real samples and $K$ is the number of possible classes of labeled data, i.e., $p_D(y=K+1|x)$ is the probability of a sample generated from $G$. Then the loss function for training $D$ comes from three parts, labeled data $\mathcal{L}$, unlabeled data $\mathcal{U}$ and data from $G$:
\begin{equation}
    \begin{aligned}
        & L_{\mathcal{L}} = -\mathbb{E}_{\bm{x}, y \sim \mathcal{L}} \:  [\log p_D(y|\bm{x}, y < K+1)], \\
        & L_{\mathcal{U}} = -\mathbb{E}_{\bm{x} \sim \mathcal{U}} \: [\log p_D(y \leq K| \bm{x})], \\
        & L_{\mathcal{G}} = -\mathbb{E}_{\bm{x} \sim G} \: [\log p_D(y = K+1| \bm{x})],
    \end{aligned}
    \label{eq:loss}
\end{equation}
And the total discriminator loss becomes $L_D = L_{\mathcal{L}} + L_{\mathcal{U}} + L_{\mathcal{G}}$.
The first term in Equation \ref{eq:loss} is the standard supervised cross-entropy loss, which minimizes the negative log probability of the label, given the data sample is labeled. The second term minimizes the negative log probability of an unlabeled sample coming from one of $K$ possible classes. The third term minimizes the negative probability of a fake sample being recognized. 

One thing to note is that the discriminator with $K+1$ outputs is over-parameterized, since the outputs of a softmax function sum to one. Thus, we can set $D$ with $K$ outputs and the equivalent discriminator is given by $\displaystyle D(x) = \frac{Z(x)}{Z(x)+1}$, where $Z(x) = \sum_{k=1}^K \exp[l_k(x)]$.

In our experiments, we set $D$ and $G$ to have the same architecture but mirroring each other, with five hidden layers. A tanh layer is added at the end of $G$ that maps the output to the range between -1 and 1 (same as input features). We used weight normalization \cite{salimans2016weight} and drop out \cite{srivastava2014dropout} to accelerate training and prevent overfitting. 

\subsection{Training and Inference}
Training GANs is notoriously difficult, particularly in a semi-supervised learning setting that requires jointly learning from labeled and unlabeled data. As noted in \cite{salimans2016improved}, using feature matching loss for $G$ works well empirically for semi-supervised learning. The objective of feature matching is guiding the generator to generate samples that match the first order statistics of real data. Furthermore, instead of directly minimizing the distance between generated sample mean and real sample mean, the discriminator was used as a feature extractor and the intermediate layer output was used as the "feature" of data samples. The loss term of feature matching is expressed as 
\begin{equation}
    L_{FM} = ||\mathbb{E}_{\bm{x} \sim G}f(\bm{x}) - \mathbb{E}_{\bm{x}\sim \mathcal{U}}f(\bm{x})||^2.
\end{equation}

A more in-depth discussion of using GANs in semi-supervised learning setting can be found in \cite{dai2017good}. It was suggested that the generator in 
SSL should generate samples that are complement to real samples. Intuitively, only if the generated sample distribution does not interfere the true sample distributions, it can help the discriminator to learn the manifolds of real samples from different classes. To achieve this, the paper proposed to increase the diversity of generated samples by increasing generator entropy, via introducing a new loss term pull-away term (PT) first proposed in \cite{zhao2016energy}:
\begin{equation}
    L_{PT} = \frac{1}{N(N-1)}\sum_{i=1}^N \sum_{j \neq i}\Big(\frac{f(x_i)^T f(x_j)}{||f(x_i)|| ||f(x_j)||}\Big)^2.
\end{equation}

Additionally, in order for complement generator to work, the discriminator needs to have strong belief on fake-real on unlabeled data. This is achieved by adding a conditional entropy loss to discriminator:
\begin{equation}
    L_{ent} = \mathbb{E}_{\bm{x}\sim \mathcal{U}}\sum_{k=1}^K p_D(k|x)\log p_D(k|x).
\end{equation}

Finally, the SSL GAN model has discriminator loss $L_D=L_{\mathcal{L}} + L_{\mathcal{U}} + L_{\mathcal{G}} + L_{ent}$ and generator loss $L_G = L_{FM} + L_{PT}$.

\section{Experiment}

\subsection{Data}

We leverage data from IQVIA longitudinal prescription (Rx) and medical claims (Dx) databases, which include hundreds of millions patients' clinical records. In our study, we focus on one type of rare disease, exocrine pancreatic insufficiency (EPI).

The detailed data preparation process is as follows. We pulled the diagnoses, procedures and prescriptions at transaction level from January 1, 2010 to July 31, 2017. We only kept a subset of patients by applying standard patient eligibility rules, which left us with a total number of 1,792,760 patients. Out of all the patients, 29,149 of them (1.6\%) are found to be diagnosed with EPI, which are labeled as positive. 80\% of the positive patients were used for training and validation and the rest were held for testing. It is important to note that the remaining patients are under-diagnosed, not essentially negative, so we cannot simply label them as is. Therefore, we applied business rules and identified 69,845 negative patients (three times as the number of positive training patients) for training and validation. The final numbers are shown in Table \ref{tab:population}. 
\vspace{-1pt}
\begin{table}[h!]
    \centering
    \begin{tabular}{c|c|c|c}
     &  Positive &  Negative & Unlabeled\\
     \hline
    Total & 29,149 & 506,450 & 1,257,161 \\
     \hline
    Train/validation & 23,395 & 69,845 & 1,257,161 \\
     \hline
    Test & 5,754 & 436,605 & 0
\end{tabular}
    \caption{Population distribution.}
    \label{tab:population}
\end{table}
\vspace{-2pt}

\subsection{Baseline}
For comparison, we chose logistic regression (LR), random forest (RF), XGBoost (XGB) and the discriminator (DNN) in the GAN architecture as baseline models. Note that the input to the benchmark classifiers is the output of LSTM embedding. 

\subsection{Evaluation Strategy}
 We used Adam optimizer \cite{kingma2014adam} to train each model, with the default learning rate set to 0.001. The number of training epoches was 20. The model was implemented and tested in Tensorflow with GPU support on a system equipped with 128GB RAM, 8 Intel Xeon E5-2683 at 2.10GHz CPUs and one Tesla P100-PCIE GPU.

Because of the high imbalance of data, we used precision-recall (PR) curve and area of PR curve (PR AUC) as evaluation metrics. The PR AUC is computed by trapezoidal rule \cite{purves1992optimum}.

%

\section{Result}
\label{sec:res}
In this section, we first present some descriptive results on medical code embedding, and then quantitative results of the model performance comparison.
\subsection{Embedding visualization}
\begin{figure}[h!]
    \centering
    \includegraphics[width=0.45\textwidth]{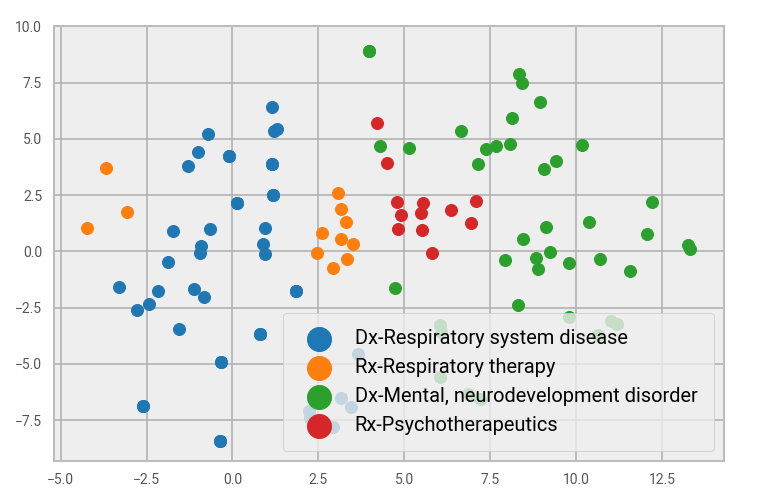}
    \caption{The visualization result by t-SNE of medical codes. Blue and orange dots are respiratory diagnosis (Dx) and prescription (Rx) codes, respectively. The green and red are Dx and Rx codes for mental diseases.}
    \label{fig:code_emb}
\end{figure}

As described in Section \ref{subsec:mce}, each medical code was represented as a 300-dimension dense vector. In order to examine whether the embedding vectors retain meaningful medical information, we identified 67 diagnosis (Dx) codes within two therapeutic areas (TAs), respiratory disease and mental disorder, as well as corresponding prescription (Rx) codes. We used t-SNE \cite{maaten2008visualizing} technique for visualization of the selected codes. The visualization result is shown in Figure \ref{fig:code_emb}. We observe that two sets of Rx codes are centered, with each forming its own cluster. The corresponding Dx codes are clustered by their TAs, and aligned with Rx codes on either side.  

\vspace{-3pt}

\subsection{Model comparison}
\begin{figure}
    \centering
    \includegraphics[width=.45\textwidth]{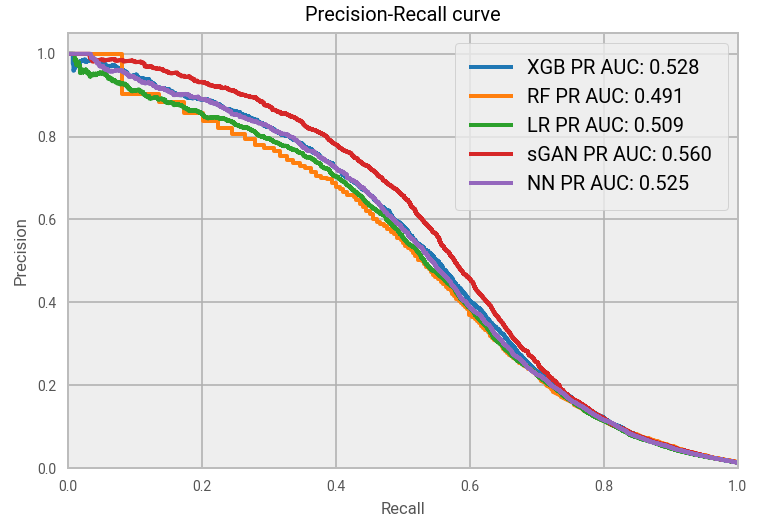}
    \caption{Precision-recall curves of the SSL GAN and benchmark models, where sGAN refers to SSL GAN model.}
    \label{fig:pr}
\end{figure}

The PR-AUC by the SSL GAN was 0.56, and the deep neural network with the same architecture as the discriminator had a score of 0.52. We saw a relative increase of 6\% over the best benchmark model. The precision-recall curves of all models are shown in Figure \ref{fig:pr}.

\section{Discussion}
\label{sec:dis}
The problem of semi-supervised learning often comes with the issue of limited labeled data, and sometimes extreme class imbalance. In our problem of interest, we had both issues. In order to improve the classification performance, it is crucial to fully make use of unlabeled data. By comparing the PR curves, we may cautiously conclude that the performance gain over DNN was from the unlabeled data.

Although the idea of generative adversarial nets is rather straightforward and intriguing, the training process is extremely cumbersome and difficult to reach convergence. According to \cite{salimans2016improved}, the training process equals to finding a Nash equilibrium of a non-convex game with continuous, high-dimensional parameters, which may fail to converge if using gradient descent based optimization algorithm \cite{goodfellow2014distinguishability}. Therefore, carefully designed loss functions are crucial to successfully using GAN-based model.

\section{Conclusion}
\label{sec:con}
In this work, we present a novel framework which combines the merits of both recurrent neural networks and generative adversarial networks. We demonstrated that GANs used in a semi-supervised learning setting can benefit from the vast number of unlabeled data to improve prediction performance, even under an extreme data imbalance scenario. Furthermore, by utilizing RNN-based networks to directly work with patient medical sequences, we are free from extensive work of feature engineering. More importantly, our preliminary analysis of the medical code embedding shows some very interesting properties that are worth investigating in the future. Finally, this framework can be easily transferred to detecting another disease of interest.


\bibliography{ref}
\bibliographystyle{icml2019}

\end{document}